\def\year{2017}\relax
\documentclass[letterpaper]{article}
\usepackage{aaai17}
\usepackage{times}
\usepackage{helvet}
\usepackage{courier}
\usepackage{graphicx}
\begin{document}
%
\title{Tracing Linguistic Relations in Winning and Losing Sides \\ of Explicit Opposing Groups}
\author{Ceyda Sanli, Anupam Mondal\\
	Rolls-Royce@NTU Corporate Lab \\
	Nanyang Technological University \\
	{\tt \{ceyda,manupam\}@ntu.edu.sg}
	\\\And
    Erik Cambria \\
	School of Computer Science and Engineering \\
	Nanyang Technological University \\
	{\tt cambria@ntu.edu.sg}}
	

\maketitle
\begin{abstract}
Linguistic relations in oral conversations present how opinions are constructed and developed in a restricted time. The relations bond ideas, arguments, thoughts, and feelings, reshape them during a speech, and finally build knowledge out of all information provided in the conversation. Speakers share a common interest to discuss. It is expected that each speaker’s reply includes duplicated forms of words from previous speakers. However, linguistic adaptation is observed and evolves in a more complex path than just transferring slightly modified versions of common concepts. A conversation aiming a benefit at the end shows an emergent cooperation inducing the adaptation. Not only cooperation, but also competition drives the adaptation or an opposite scenario and one can capture the dynamic process by tracking how the concepts are linguistically linked. To uncover salient complex dynamic events in verbal communications, we attempt to discover self-organized linguistic relations hidden in a conversation with explicitly stated winners and losers. We examine open access data of the United States Supreme Court. Our understanding is crucial in big data research to guide how transition states in opinion mining and decision-making should be modeled and how this required knowledge to guide the model should be pinpointed, by filtering large amount of data.
\end{abstract}

\section{Introduction}

Traditionally, in computational linguistics, it is essential to integrate models and algorithms with fundamental laws of language. Widely applied hierarchical dependency trees and parsing in natural language processing (NLP) follow existing grammatical relations. Nowadays, while algorithms and models reach higher levels and available data becomes bigger, not enough linguistic laws are uncovered and can have a chance to meet with developed techniques. Language processing in data science mainly considers evaluated data as single source in terms of language. There are approaches such as cross-media topic analysis, retrieving information referring various data platforms including websites, blogs, and mobile phones, and multimodal analysis~\cite{porcon,Poria201798_RevMultiModal,porens}, combining text data with images, videos, and audio, however, they only gather all available channels and do not address the richness of language. 

On the other hand, language itself has many dimensions, language of a text written by a single author is different than language used in a dialogue or that of a group speech, e.g., trialogue discussions. Therefore, it is emergent that current conventional NLP should meet with the revolutionary philosophy of linguistics~\cite{Chomsky_1975_Reflections_on_Language} and establish new hidden laws applicable in data science: the human mind easily knows and applies by birth, but hardly formulates to understand the underlying structure. 

One of the remarkable perspectives to dig into natural linguistic laws is provided by social and behavior sciences, adaptation in language during communication as a result of changes in opinions and decisions. Opinions and decisions are personal in individual level, however, they are flexible while facing public opinions and decisions. Linguistic adaptation is twofold. In one part, collective voice unifies opinions and decisions in a complex process, ideas are biased, and consequently people start acting similarly, talking similarly, and so writing similarly. Twitter conversations~\cite{DanescuCristian2011_MarkMyWords,Purohit20132438} and popular memes~\cite{ClashofContagions2012,CompetitionSuccessMeme2013} prove this similarity in social media. 

In the other part, when people have a well-defined goal at the end, they tend to reshape their arguments. In the presence of distinct winning and losing sides and social hierarchy, people at lower status show both cooperation through that at the higher status and competition among each other. Therefore, a verbal discussion in such explicitly opposing groups host linguistic adaptation, investigated in social exchange theory~\cite{ExchangeTheory,FromStatustoPower2006}. 
While information and emotions are the fundamental elements of human knowledge, commonsense knowledge is the fundamental element for gluing society \cite{Cambria2009_CommonSenseComputing,camnt4}. Commonsense is implicit semantic and affective information humans continuously tap on for decision-making, communication, and reasoning in general~\cite{camsen,dhegra,porcom,tratow}. Effective speeches and public talks use commonsense efficiently to drive opinions and change decisions in large scales~\cite{MakingCommonSense1994_Leadership}. The resultant unified collective motion is extremely interesting in social groups~\cite{spanishprotestPLoSONE,spanishprotestSciRep}. 

Opinions and decisions are personal in individual level. However, as observed, they are quite flexible facing with a collective decision. Complex knowledge extraction process in micro state suddenly becomes less valuable and group decision gains~\cite{PoliticalPolarizationonTwitter2011}. We can argue that our opinions are biased when our decisions mostly rely on our previous knowledge, e.g., commonsense, and so richness of opinions kept in each individual is relatively unimportant. We can further argue that commonsense drives an adaptation in extracting knowledge. To measure commonsense for a particular situation is hard, however, adaptations can be easily captured in Twitter conversations~\cite{DanescuCristian2011_MarkMyWords,Purohit20132438}, in memes~\cite{ClashofContagions2012,CompetitionSuccessMeme2013}, and face-to-face discussions~\cite{DanescuCristian2012_EchoesofPower}.

In this paper, our main concerns are firstly to construct discussion groups including agents having different social powers and serving opposite aims. Secondly, we investigate how we can track the progress of opinions together with their influences on decisions in oral conversations. We claim that linguistic relations~\cite{Poria_2015_DynamicLinguisticPatterns} preserve all rich phenomena, shortly discussed above, including collective voice, reshaping arguments, and so adaptation. To analyze adaptation induced by both cooperation and competition, we consider court conversations: they are held in clearly stated winner and loser groups with distinct hierarchy in decision-making due to the presence of Justices and lawyers. 

To this end, we evaluate the open access data of the United States Supreme Court~\cite{Hawes2009_SupremeCourt,Hawes_MSThesis_SupremeCourt,DanescuCristian2012_EchoesofPower}, prepare conversation groups with different adaptation levels, implement a suitable algorithm to extract linguistic relations in these group conversations, and finally provide a comparison between the groups and the discovered linguistic relations.

The rest of the paper is organized as follows: the first section presents the dataset we consider and designed conversation groups out of the data; the second section describes our algorithm in detail; the following section explains how we implement pointwise mutual information for the conversation groups and then link with linguistic relations; finally, we provide experimental results and conclude the paper.

\section{Supreme Court Data}

We borrow the textual data of the conversations in the United States Supreme Court pre-processed by~\cite{Hawes2009_SupremeCourt,Hawes_MSThesis_SupremeCourt} and enriched by~\cite{DanescuCristian2012_EchoesofPower} including the final votes of Justices. Both the original data and the most updated version used here are publicly available~\cite{DanescuCristian2012_EchoesofPower}. The data gathers oral speeches before the Supreme Court and hosts 50,389 conversational exchanges among Justices and lawyers. 

Distinct hierarchy between Justices (high power) and lawyers (low power) impose lawyers to tune their arguments under the perspective and understandings of Justices, and as a result, speech adaptation and linguistic coordination leaves their traces in a sudden occurrence of sharing the same adverbs, conjunctions, and pronouns. Tracking initial utterances, the sides present a unique and personal speaking, but after a while in the communication, word selections, their forms, and frequencies mirror each other's language preference. The linguistic coordination is systematically quantified by~\cite{DanescuCristian2012_EchoesofPower} and the arguments follow the principles of exchange theory examining behavior dynamics in low and high power groups~\cite{ExchangeTheory,FromStatustoPower2006}: Lawyers tend to cooperate more to Justices than conversely and demonstrate strong linguistic coordination in their speech. Moreover, lawyers show even more cooperation to unfavorable Justices than favorable ones.

Here, we enrich the comparison including the identity of winners and losers in lawsuits. The data provides whether the petitioner or the respondent is the winner at the end of each lawsuit. In addition, the speaker of each utterance is labeled as their position, e.g., Justice or lawyer. Furthermore, Justice's votes and the side of lawyers are tagged with the utterances.  Table~\ref{table:Positions_JusticeLawyers} identifies all roles carried by Justices and lawyers. For Justices, both the vote (middle) and whom to speak (last) are given. Lawyers are allowed to speak only when Justices address their side. 

\begin{table}[ht!]
\begin{tabular}{l l c c}
ID & Roles of Justices ($J$) and Lawyers ($l$) \\
\hline
1 & $J$ - Vote Petitioner - Speak to Petitioner's $l$ \\
2 & $J$ - Vote Petitioner - Speak to Respondent's $l$ \\
3 & $J$ - Vote Respondent - Speak to Petitioner's $l$ \\
4 & $J$ - Vote Respondent - Speak to Respondent's $l$ \\
5 & $l$ - Petitioner Side\\
6 & $l$ - Respondent Side\\
\end{tabular}
\caption[Table caption text]{The segregation schema of the roles in conversations: Support sides of Justices and sides of lawyers. 1-6 summarize all potential roles present in the data. In 1-4, who supported by the Justice is given in the middle. Furthermore, the last indicates the side of lawyer the Justice speaks to.}
\label{table:Positions_JusticeLawyers}
\end{table}

Referring exchange theory~\cite{ExchangeTheory,FromStatustoPower2006} and the measured coordination~\cite{DanescuCristian2012_EchoesofPower}, one can order the relative power of each Justice and lawyer pair
\begin{equation}\label{Eq:Power_basic}
P(J_u, l) > P(J_s, l),
\end{equation}
where $J$ and $l$ represent Justices and lawyers, respectively (note that for comparing individually following the social exchange theory, $P(J)>P(l)$ for both supportive and unsupported Justices). The subscript $u$ indicates that Justice doesn't support the side of lawyer and the supportive version is described by $s$. For instance, in Table~\ref{table:Positions_JusticeLawyers}, in the communications of 1 and 5; 4 and 6, Justices show supports and play as $J_s$, whereas that of 3 and 5; 2 and 6, lawyers are unsupported by $J_u$. The scenarios and pairs guide to construct groups with different cooperation level induced by $P$ as illustrated in Table~\ref{table:Grouping_by_Coordination}. 

\begin{table}[ht!]
\begin{tabular}{l l l c c c}
Group ID & Cooperation & Pool of $J$ and $l$\\
\hline
I.i & supportive, $P(J_s, l)$ & 1 and 5 \\
I.ii & unsupported, $P(J_u, l)$ & 3 and 5 \\
II.i & unsupported, $P(J_u, l)$ & 2 and 6 \\
II.ii & supportive, $P(J_s, l)$ & 4 and 6 \\
\end{tabular}
\caption[Table caption text]{Grouping communications with respect to level of cooperation, based on the relative power of the partners in the conversations. 1-6 and the power pairs $P(J_s, l)$ and $P(J_u, l)$ as defined previously.}
\label{table:Grouping_by_Coordination}
\end{table}

We further add another dimension in the relative power: Winners and Losers, haven't been investigated in the previous study~\cite{DanescuCristian2012_EchoesofPower}. To this end, Eq.~\ref{Eq:Power_basic} is reformulated
\begin{eqnarray}
P(J_u, l)_{\mbox{\scriptsize win}} &>& P(J_s, l)_{\mbox{\scriptsize win}}, \\ 
P(J_u, l)_{\mbox{\scriptsize lose}} &>& P(J_s, l)_{\mbox{\scriptsize lose}}. 
\end{eqnarray}

Here, win and lose subscripts highlight that the concerned Justice and lawyer pairs are the partners in a won or lost lawsuit. As an illustration, $P(J_s, l)_{\mbox{\scriptsize win}}$ occurs in the group I.i when petitioners are the winner and also in II.ii while respondents are the winners of the lawsuits. On the other hand, $P(J_s, l)_{\mbox{\scriptsize lose}}$ is the Justices-lawyers of I.i in respondent won lawsuits as well as of II.ii in petitioner won lawsuits. The situations are generated for the unsupported Justice-lawyer groups and all are listed in Table~\ref{table:Final_Conversation_Groups}. 

\begin{table}[ht!]
\begin{tabular}{l l l c c c}
$\kappa$ & Cooperation & Gathering Group ID  \\
\hline
$A$ & supportive, win: $P(J_s, l)_{\mbox{\scriptsize win}}$ & I.i of Pe + II.ii of Re \\
$B$ & supportive, lose: $P(J_s, l)_{\mbox{\scriptsize lose}}$ & I.i of Re + II.ii of Pe \\
$C$ & unsupported, win: $P(J_u, l)_{\mbox{\scriptsize win}}$ & I.ii of Pe + II.i of Re \\
$D$ & unsupported, lose: $P(J_u, l)_{\mbox{\scriptsize lose}}$ & I.ii of Re + II.i of Pe \\
\end{tabular}
\caption[Table caption text]{Designed conversation groups $\kappa$ based on different expectations for the level of linguistic coordination, induced by distinct $P$. The groups $\kappa$ are presented in $A$, $B$, $C$, and $D$, whether they preserve supportive or unsupported conversations as well as a winner or loser status stated by the Supreme Court. Pe and Re represent the particular lawsuits where Petitioner and Respondent as the winner, respectively. Gathered conversations of the cases I.i, I.ii, II.i, and II.ii and the relative powers $P$ are as introduced earlier.}
\label{table:Final_Conversation_Groups}
\end{table}

Calculating utterances in $\kappa$, we have 21,105 for $A$, 15,116 for $B$, 15,489 for $C$, and 24,461 for $D$, gathered by different combinations of 195 lawsuits. The large number of each pool convinces that we have enough examples to perform statistics and our measurement won't be biased by the size effect. 
On the other hand, noting the total number of 50,389 utterances, almost the half of the data presents $P(J_u, l)_{\mbox{\scriptsize lose}}$ type social relations, e.g., case $D$.
Eqs. (2) and (3) do not include the comparison of $\{P(J_u, l)_{\mbox{\scriptsize win}};P(J_s, l)_{\mbox{\scriptsize lose}}\}$ and $\{P(J_u, l)_{\mbox{\scriptsize lose}};P(J_s, l)_{\mbox{\scriptsize win}}\}$ on purpose since it is unknown whether $P(J_u, l)>P(J_s, l)$ is still valid in the presence of win and lose, bringing interesting perspective while coupling the power hypothesis with the cooperation and not considered in social exchange theory. We aim to understand this full picture by correlating determined linguistic relations with the separated relative power groups.

\section{Linguistic Relation Extraction}
\begin{figure}[h]
\begin{center}
\includegraphics[width= 8.2cm]{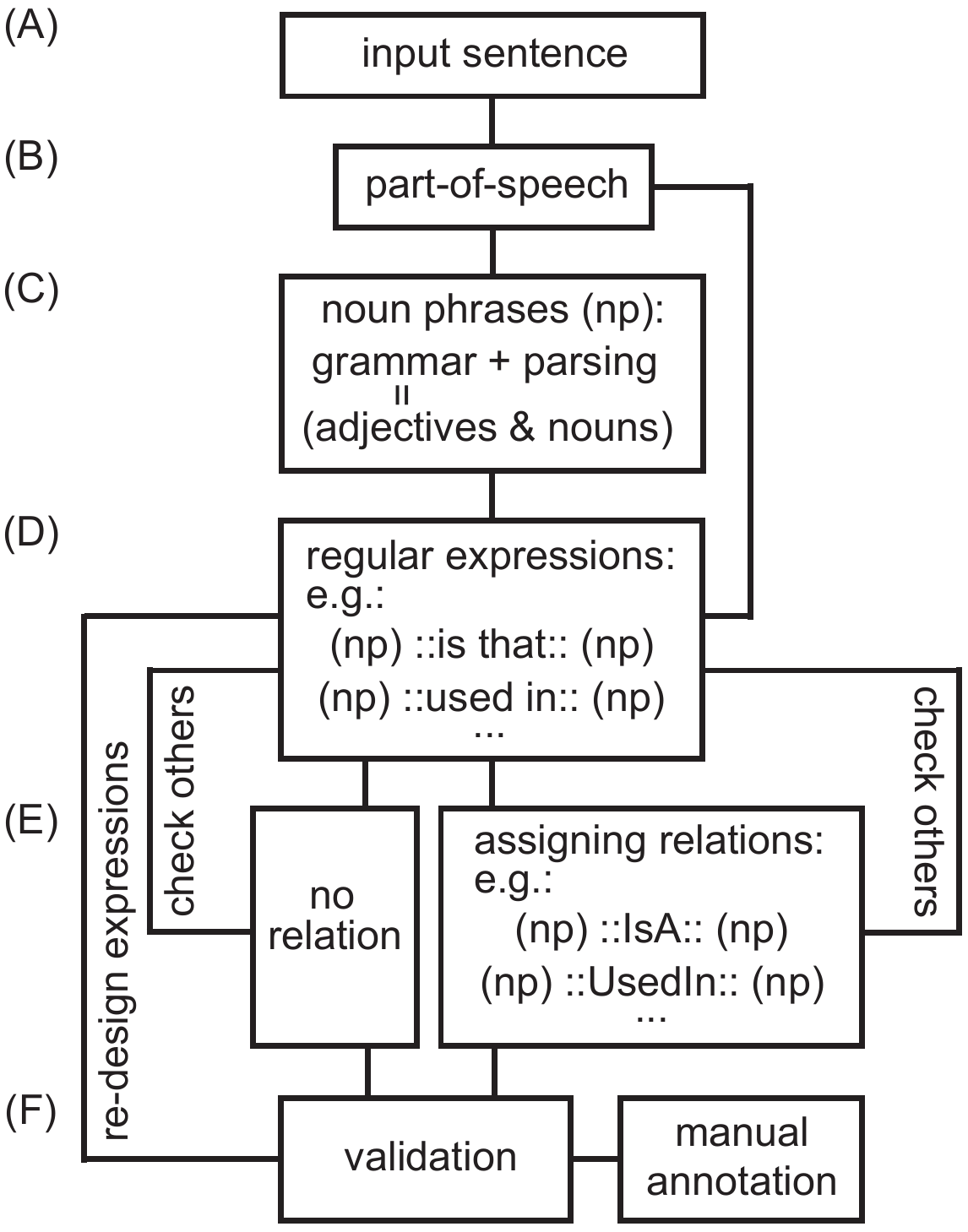}
\caption{Flow diagram of the rule-based relation extraction algorithm. The important steps are summarized from (A) to (F): (A-C) present suggesting concepts based on noun phrases of combined adjectives and nouns. (D-E) describe the iteration of applying designed static surface patterns (regular expressions) together with supervising for the 6 relations, namely, IsA, PartOf, UsedBy, UsedFor, UsedIn, UsedOver, and UsedWith. (F) indicates the final step of validation compared with the manual annotation set and formulating again regular expressions in (D) to increase the performance.
\label{fig:Fig1_RE_FlowDiagram}}
\end{center}
\end{figure}

The Supreme Court hosts lawsuits of rich subjects. To design specific linguistic relations in each distinct lawsuit is challenging and not required. 
Our aim is to suggest relations suitable for any discussion concept. To generalize the task, we first determine noun phrases in the data following the definition in~\cite{Espresso_AutomaticHarvesting_2006}. The phrases are combinations of adjectives and nouns. The technical steps include standard part-of-speech tagging including grammar based chunk parser. We then restrict our attention to address the relations linking only determined noun phrases within one sentence. The data shows utterances of grammatically correct and well-organized sentences. To this end, we apply rule-based relation extraction. While Fig.~\ref{fig:Fig1_RE_FlowDiagram} shows each step of the developed algorithm, steps (A-C) indicate the discussed concept recognition of noun phrases.

The rule-based schema starts with first restricting linguistic relations and then constructing static surface patterns (regular expressions) for them. The assigned patterns run as an iterative process searching the exact match of the real patterns between any concept pair, which is any noun phrase pair here. Within a sentence, multiple relations can be addressed based on the comparison in the iteration, to capture both different relations or the same relation but with the different patterns. To balance the relations without overweighting extreme cases, we first apply classical IsA~\cite{IsA_Hearst1992} and PartOf~\cite{PartOf_Girju2003} relations. The patterns of the relations follow both lexico-syntactic formalisms~\cite{LexicoSyntacticPatterns_Ontology_Klaussner2011} and manual investigations of the data.

We then recommend further relations as UsedBy, UsedFor, UsedIn, UsedOver, and UsedWith to cover the rest of the data. The Used relations do not accumulate for certain lawsuits and nicely distribute over entire data, which provides us reliable analysis. Fig.~\ref{fig:Fig1_RE_FlowDiagram}(D and E) highlight the iteration process to detect all potential relations.
To illustrate the outcome of our algorithm, we provide examples for each relation. They are given with the detected noun phrases in Table~\ref{table:Example_RelationExtraction}. The identity of the sentences, a-g, are to guide the following concerned examples, where the linked noun phrases are highlighted in bold: \\
\noindent (a) That was so because her \textbf{claim} is that J.$\_$Howard intended to give her a \textbf{catchall$\_$trust}.\\
\noindent (b) And when you look at the \textbf{core$\_$value} of the two \textbf{clauses}, they do not clash.\\
\noindent (c) And what I'm trying to do here for the Court is to draw upon your own$\_$authority, the word you've spoken, as opposed to the \textbf{test} proposed by the \textbf{Criminal$\_$Justice$\_$Foundation} and by the \textbf{United$\_$States}. \\
\noindent (d) One, the manufacturing$\_$process allows there to be a \textbf{safe$\_$use} for one of the \textbf{components }in \textbf{marijuana}. \\
\noindent (e) The \textbf{phrase$\_$Justice$\_$Harlan} used in the \textbf{Davis$\_$case}.\\
\noindent (f) For 124 years, as \textbf{state$\_$power} over \textbf{alcohol} has ebbed and flowed. \\
\noindent (g) The \textbf{haulers} are required today to comply with the \textbf{program}.

\begin{table}[ht!]
\begin{tabular}{l c c c c c}
Relation & Linked Noun Phrases & Sent. ID \\
\hline
IsA & claim :: catchall$\_$trust & (a) \\
PartOf & core$\_$value :: clauses  & (b) \\
UsedBy & test :: Criminal$\_$Justice$\_$Foundation,  & (c) \\
 & United$\_$States & \\
UsedFor & safe$\_$use :: components, & (d)\\
 & marijuana  &  \\
UsedIn & phrase$\_$Justice$\_$Harlan :: Davis$\_$case  & (e) \\
UsedOver & state$\_$power :: alcohol & (f) \\
UsedWith & haulers :: program & (g) \\
\end{tabular}
\caption[Table caption text]{Extracted relations with our algorithms and the corresponded (linked) noun phrases. Sent. ID refers the labeled example sentences above in the main text.}
\label{table:Example_RelationExtraction}
\end{table}

The validation of the discovered linguistic relations and their suggested patterns are systematically satisfied by the following protocol. From each conversation group $\kappa$ in Table~\ref{table:Final_Conversation_Groups}, 1000 utterances are randomly selected. Utterances present averages sentences of 2-4, the minimum is for the group $C$, $P(J_u, l)_{\mbox{\scriptsize win}}$, and the maximum for group $D$, $P(J_u, l)_{\mbox{\scriptsize lose}}$. 

Then, manual annotations are provided for each pool, which works as the ground truth, and the patterns are re-adjusted if necessary based on the performance, as shown in Fig.~\ref{fig:Fig1_RE_FlowDiagram}(D-F). The overall average scores, comparing the relations generated by our algorithm with the ground truth, are obtained as 59.92$\%$ for Recall, 67.2$\%$ for Precision, and 63.35$\%$ for the resultant F1. The scores are relatively higher than that of the rule-based relation extraction algorithms for more general purposes applied in large data sets~\cite{EditDistance_Pantel2004}. Our manual efforts, the grammatically correct sentences, and relatively small and well-organized data are the reasons behind the good performance. However, we observe that the foremost reason is the linguistic coordination extracting many relations from the same static patterns.

In the rest of the paper, we will demonstrate how we interpret these linguistic relations in the Supreme Court conversation groups of different relative powers.
 
\section{Pointwise Mutual Information}

Pointwise mutual information (PMI) is a metric to measure coincidence of two discrete random events. It combines individual probabilities of events and their joined probability to determine how often the two events occur at the same occasion. We quantify to what extend linguistic relations $R$ are addressed by conversation groups $\kappa$ and whether we observe any variation in the selections. 

To this end, PMI between $R$ and $\kappa$ is introduced~\cite{EditDistance_Pantel2004}
\begin{equation} \label{Eq:PMI}
MI(R, \kappa) = \log\frac{\frac{f(R, \kappa)}{N}}{\frac{\sum\limits_{R_i}^{\mbox{\tiny all $R$}}f(R_i, \kappa)}{N}\times\frac{\sum\limits_{\kappa_j}^{\mbox{\tiny all $\kappa$}}f(R, \kappa_j)}{N}}.
\end{equation}
Here, $f(R, \kappa)$ represents the frequency of occurrence for certain $R$ in particular $\kappa$ and $N$ is the total number of all $R$ in all $\kappa$. So, while the numerator describes the probabilistic occurrence of $R$ in $\kappa$, the denominator provides individual probability of $R$ and that of $\kappa$ in the pool. We expect high $MI(R, \kappa)$ while $R$ appears in a specific $\kappa$ and that is an indicator of its rare presence in the other conversation groups. 

Unlike the previous study~\cite{DanescuCristian2012_EchoesofPower}, entirely tracking back and forth utterances and proving the adaptation, e.g., linguistic coordination, by identifying the frequency of selected keywords, we directly utilize their overall conclusion and claim that linguistic relations already preserve the adaptation and any other complex collective linguistic process induced by both cooperation and competition in different power groups. We expect that the variation in $MI(R, \kappa)$ of gathered utterances of each relative power group, independent of the utterance order, suggests which relations can distinguish the difference in the groups and the magnitude of $MI(R, \kappa)$ of that difference highlights which relative power groups drastically influence the applied language. We will analyze $MI(R, \kappa)$ following this discussed understanding in coming Section.
 
\section{Results} 

\begin{figure}[ht!]
\begin{center}
\includegraphics[width= 9.2cm]{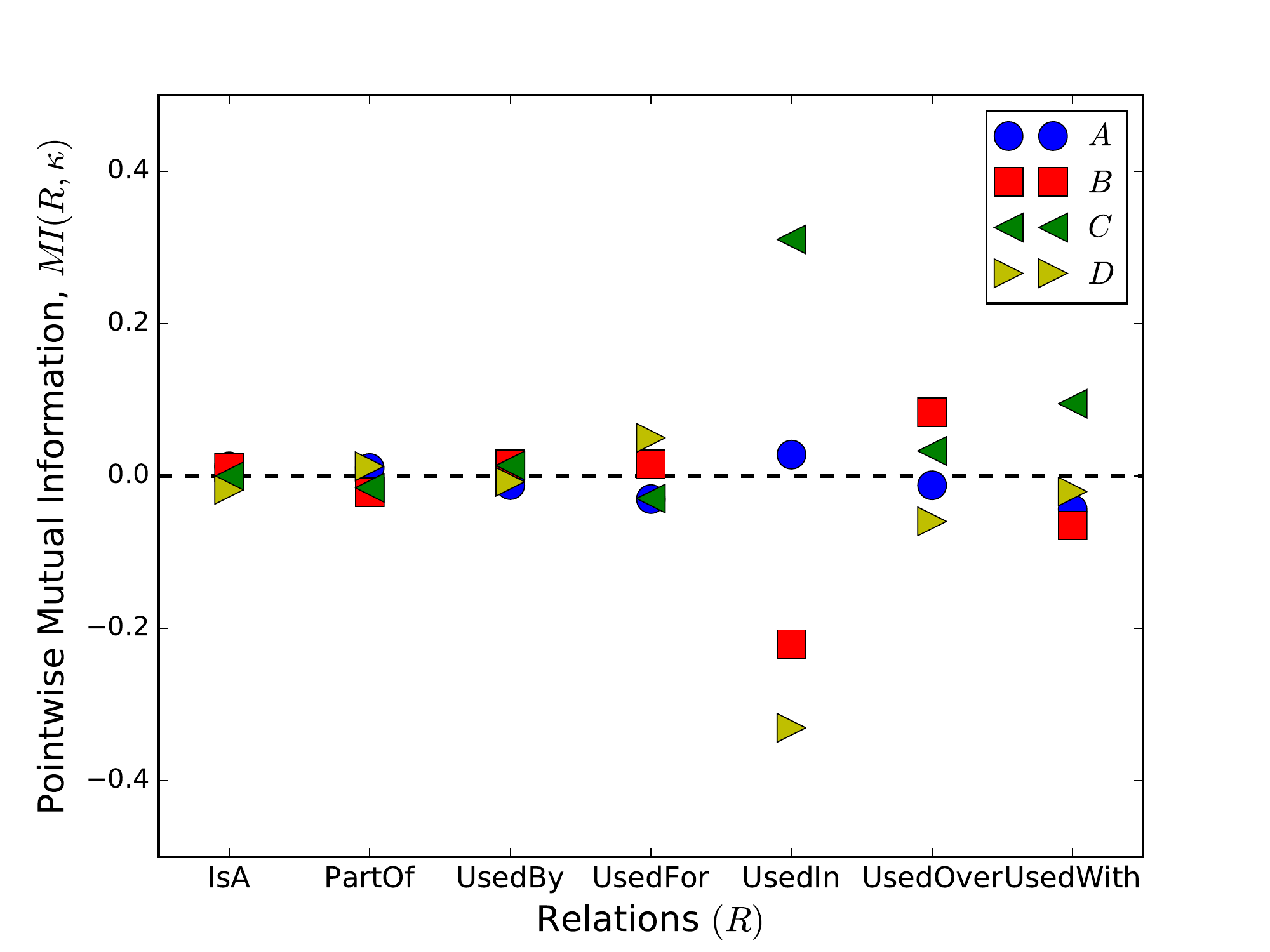}
\caption{PMI between relations $R$ and conversation groups $\kappa$: $MI(R, \kappa)$. The overall values indicate that, unlike IsA, PartOf, and UsedBy relations, the occurrences of UsedFor, UsedIn, UsedOver, and UsedWith are driven by the relative power and the resultant linguistic coordination and further complex process. The marker representations are as follows: Circles (blue) for $A$, squares (red) for $B$, left triangles (green) for $C$, and right triangles (yellow) for $D$.
\label{fig:Fig2_MI}}
\end{center}
\end{figure}

We perform $MI(R, \kappa)$ for each group $\kappa$ separated by different coordination level and linguistic dynamics, expected due to the distinct relative powers as introduced in Table~\ref{table:Final_Conversation_Groups}, and each relation $R$ described in Section Linguistic Relation Extraction. The results are presented in Fig.~\ref{fig:Fig2_MI} and suggests rich behavior. First, $MI$ for the relations IsA, PartOf, and UsedBy is almost indistinguishable overall $\kappa$. We understand that these relations cannot uncover the linguistic variations in different power groups. This is an obvious outcome of NLP and examining sentences by lexico-syntactic patterns: Any sentence can consider them with no complex linguistic process such as coordination and competition. On the other hand, we observe quite remarkable separation starting with UsedFor. Successfully, the results of UsedIn, UsedOver, and UsedWith show that their appearances in $\kappa$ are not arbitrary.

Evaluating the results in more detail, let us remind Table~\ref{table:Final_Conversation_Groups}. $A$ is expected to have the least relative power, $P(J_s, l)_{\mbox{\scriptsize win}}$, and consequently, no significant variation is observed. However, the situations are much more challenging for $B$, $C$, and $D$: They face with many conceptual challenges while defending their sides and competing with the opposite arguments, $C$ and $D$, and to experiment different communications in a losing state, $B$ and $D$. Each difficulty is a potential origin of the competition, some can build sufficient cooperation and make the lawyer winner, $C$, some cannot help to overcome the situation, keep the coordination limited, and so we experience lost lawsuits, $B$ and $D$. 
To remind, $B$ for $P(J_s, l)_{\mbox{\scriptsize lose}}$, $C$ for $P(J_u, l)_{\mbox{\scriptsize win}}$, and $D$ for $P(J_u, l)_{\mbox{\scriptsize lose}}$. 
If we just call social exchange theory, for any measurable linguistic quantity, we would need to have $A \equiv B$ and $C \equiv D$. However, we show that the win and lose states impose observable deviations and none group resembles each other, oppositely, each presents very unique behavior. In a simplified picture, $MI(R, \kappa)$ for $C$ always indicates significantly positive values. This proves that the utterances in $C$ consider all type of relations, can be the reason behind the success of the ``win" state in spite of the presence of unsupported Justices.  

\section{Conclusion}

We investigated the linguistic dynamics in terms of a restricted set of linguistic relations in oral conversations while the actors have different powers such as Justices (high power) and lawyers (low power) in the United States Supreme Court. Initially, defined cooperation of lawyers to Justices and the resultant linguistic coordination are only based on the relative power. This is a microscopic picture underestimating the dynamics of emergent competition arises in a losing state (lost lawsuits), which can change the nature of the linguistic coordination and make the linguistic relations richer. Our argument is proven by measuring $MI(R, \kappa)$ always positive for the group $C$, $P(J_u, l)_{\mbox{\scriptsize win}}$. Novelty of our approach is that it evaluates supportive and unsupported situations in more realistically. The principle of exchange theory suggests $P(J_u, l)>P(J_s, l)$ and one should expect high coordination in the former.

However, this can be always true if there is no explicitly stated decision at the end of the communication: Winner or loser lawyer. We can observe $P(J_s, l)_{\mbox{\scriptsize lose}} \simeq P(J_u, l)_{\mbox{\scriptsize lose}}$ and so the linguistic coordination (dynamics) for both can be comparable, as we trace in our result, e.g., very similar trend of $MI(R, \kappa)$ for groups $B$ and $D$. Therefore, both social exchange theory and their impacts on the linguistic behavior need to be reinterpreted under exogenous factors such as win-lose situations. 
Furthermore, we experience that the rule-based relation extraction is well-applicable for speech data, in this grammatically correct form with minor noise, because of the presence of the linguistic adaptation, providing a better performance than its usage for other type of textual data such as internet data. Furthermore, $MI(R, \kappa)$ brings another perspective to uncover complex linguistic dynamics, including cooperation and competition, and discover the correlations between the linguistic relations and the relative powers. 
We establish the preliminary set-up to examine the linguistic dynamics of trialogue discussions hosting in social groups with distinct hierarchy. 

Our main conclusion is that win and lose states impose further complexity and change the conventional application of social exchange theory in language and communication. In our future study, we attempt to analyze back and forth utterances in detail regarding semantics bonding by the linguistic relations by applying advanced tools. 

\section{Acknowledgments}
This work was conducted within the Rolls-Royce@NTU Corporate Lab with support from the National Research
Foundation (NRF) Singapore under the Corp Lab@University Scheme. We thank San Linn for his useful comments on the rule based relation extraction approach.

\bibliography{refs_TracingLinguisticRelations}
\bibliographystyle{aaai} 
\end{document}